\documentclass[11pt]{article}
\usepackage[T1]{fontenc}
\usepackage{amssymb}
\usepackage{amsmath}
\usepackage{graphicx}
\usepackage{authblk}
\usepackage[letterpaper, margin=1in]{geometry}
\usepackage{tikz}
\usetikzlibrary{positioning, arrows.meta, fit, calc, backgrounds}

\date{}
\title{Multimodal Molecular Representation Learning with Graph Neural Networks, Deep \& Cross Networks, and SMILES Embeddings}
\author{
	Qiwei Han\textsuperscript{1,2,$\ast$}, 
	Chi Zhou\textsuperscript{2,$\ast$},
	Ruobing Wang\textsuperscript{1,2},
	Zheng Ma\textsuperscript{1,2}
	
	\vspace{0.2cm} % Adds a small gap before the affiliations
	
	\small{\textsuperscript{1}\textit{Department of Chemistry, Duke University, Durham, NC 27708, USA}} \\
	\small{\textsuperscript{2}\textit{Department of Computer Science, Georgia Institute of Technology, Atlanta, GA 30332, USA}} \\
	
	\vspace{0.1cm} % Adds a small gap before the footnote
	
	\small{\textsuperscript{$\ast$}\textit{These authors contributed equally to this work.}}
}
\begin{document}
	\maketitle
	
\begin{abstract}
	Molecular property prediction often relies on isolated data modalities, where continuous 3D graph neural networks (GNNs) struggle to efficiently capture long-range topological dependencies and exact macroscopic heuristics. In this work, we introduce a parameter-efficient Tri-Branch Modular Fusion Neural Network that synthesizes three orthogonal modalities: 3D spatial geometry (SchNet), discrete topological grammar (SMILES via ChemBERTa), and explicit macroscopic physicochemical descriptors (Deep \& Cross Network). By bypassing standard scalar readouts and employing a shared late-fusion architecture, the framework establishes a mathematically rigorous multimodal latent space that effectively resolves the arithmetic and oversmoothing limitations of local message passing. We evaluate the proposed architecture on the QM9 benchmark, targeting the extensive thermodynamic property of atomization energy at 0 K ($U_0^{\mathrm{atom}}$). Through systematic combinatorial ablation and latent bottleneck optimization ($d_e=64$), the tri-modal framework achieves a validation Mean Absolute Error (MAE) of 0.0207 eV. Operating with fewer than one million parameters, this architecture decisively surpasses the sub-chemical accuracy threshold and yields a substantial 20.6\% error reduction over a strictly controlled geometric baseline. Ultimately, our findings demonstrate that integrating orthogonal macroscopic and topological data streams provides a synergistic, $\mathcal{O}(1)$ physical shortcut. This multimodal alignment offers a highly efficient alternative to brute-force parameter scaling, establishing a robust surrogate model for high-throughput virtual screening (HTVS) pipelines.
\end{abstract}


\section{Introduction}

The accurate prediction of molecular properties is a fundamental task in computational chemistry, drug discovery, and materials science. Properties such as atomization energy, dipole moments, and electronic characteristics are traditionally calculated using quantum mechanical methods like Density Functional Theory (DFT) \cite{kohn1965self}. While DFT provides highly reliable predictions, its immense computational cost strictly limits its applicability in large-scale workflows. Consequently, machine learning (ML) has emerged as a critical surrogate, approximating DFT-derived properties at a fraction of the computational expense.

The development of molecular ML has been accelerated by standardized benchmarks like MoleculeNet \cite{wu2018molecule} and generative models that map discrete representations to continuous latent spaces \cite{gomez2018automatic}. Recently, geometric deep learning has dominated this space by learning directly from 3D spatial structures. Foundational models like SchNet \cite{schutt2017schnet} demonstrated the power of continuous-filter convolutions, while subsequent equivariant networks have achieved remarkable error floors on standard datasets like QM9 \cite{ramakrishnan2014quantum}. 

However, this pursuit of absolute precision has driven the field toward heavily parameterized architectures. Relying on complex high-order tensors or massive attention mechanisms, state-of-the-art equivariant models \cite{satorras2021n, batatia2022mace} often require millions of parameters. While these architectures achieve exceptional absolute accuracy, their computational overhead and high VRAM requirements present a bottleneck for real-world applications like high-throughput virtual screening (HTVS). In HTVS pipelines, where chemical libraries frequently exceed millions of candidate compounds, rapid inference and parameter efficiency are as critical as sub-chemical accuracy. This dynamic highlights a pressing need for alternative paradigms: achieving robust predictive fidelity not through the brute-force parameter scaling of a single spatial modality, but through the computationally lightweight synthesis of complementary data sources.

Monolithic, geometry-only architectures often ignore the heterogeneous nature of chemical data. Beyond 3D spatial coordinates, molecules are richly characterized by explicitly calculated physicochemical descriptors (0D macroscopic scalars) and structural sequences (1D SMILES strings) \cite{weininger1988smiles}. While modern message-passing GNNs excel at resolving localized quantum micro-environments, they often struggle with exact global counting tasks and long-range topological dependencies due to graph oversmoothing. Conversely, tabular architectures (like Deep \& Cross Networks) can easily map explicit macroscopic heuristics, and language models excel at capturing long-range topological grammar via global self-attention. Combining these modalities theoretically provides an orthogonal mathematical shortcut that mitigates the blind spots of GNNs, though it remains a non-trivial challenge due to stark differences in feature distributions and information density \cite{wang2024multi}.

To address this, we propose a highly parameter-efficient, multimodal representation learning framework that systematically integrates continuous geometric features, discrete semantic embeddings, and explicitly bounded physicochemical descriptors. Our modular architecture processes each modality via a dedicated encoder before fusing them into a shared predictive latent space. The geometric branch captures local interatomic dependencies using continuous-filter convolutions adapted for extensive properties via additive pooling—a mathematical necessity for predicting extensive thermodynamic targets like atomization energy. Concurrently, a Deep \& Cross Network (DCN) \cite{wang2017deep} explicitly models global physicochemical heuristics via polynomial interactions. Finally, the semantic branch distills pre-trained SMILES embeddings via SwiGLU-gated dimensionality reduction \cite{shazeer2020glu}, capturing long-range functional group motifs often obscured by localized Euclidean convolutions. 

Evaluated on the QM9 benchmark for the extensive atomization energy property, our main contributions are summarized as follows:
\begin{itemize}
	\item \textbf{Parameter-Efficient Multimodal Architecture:} We propose a modular, tri-branch fusion framework integrating a spatial GNN, explicit DCN feature crossing, and SwiGLU-gated SMILES embeddings. By relying on multimodal synergy rather than extreme depth, the framework operates with fewer than one million parameters, making it highly suitable for rapid inference pipelines.
	\item \textbf{Sub-Chemical Accuracy:} The proposed fusion network achieves a validation Mean Absolute Error (MAE) of 0.0207 eV on the QM9 benchmark, successfully surpassing the standard sub-chemical accuracy threshold ($\approx$ 0.0433 eV) while maintaining a lightweight computational footprint.
	\item \textbf{Synergistic Combinatorial Ablation:} Through comprehensive ablation across unimodal and bimodal configurations, we empirically demonstrate that synthesizing orthogonal macroscopic and topological data yields a substantial 20.6\% reduction in error compared to a strictly controlled, pure geometric baseline (0.0261 eV).
	\item \textbf{Latent Bottleneck Analysis:} We provide a systematic sensitivity analysis of the semantic embedding dimensionality (sweeping from 32 to 256), identifying exactly 64 dimensions as the optimal information bottleneck required to distill global topology without over-parameterizing the downstream latent space.
	\item \textbf{Architectural Tabular Validation:} We evaluate the tabular routing mechanism, confirming that explicit polynomial feature crossing via the DCN establishes a superior validation floor compared to implicit dense mixing via a standard MLP control (0.0210 eV).
\end{itemize}

\section{Multimodal Learning Framework}

\subsection{Architecture Overview}
The proposed Modular Fusion Network is designed to synthesize three orthogonal molecular modalities via a late-fusion architecture. This paradigm ensures each data stream is processed by a structurally optimized encoder, preventing feature interference during initial representation learning. Formally, we define a molecule $M$ as a multimodal tuple:
\begin{equation}
	M = (\mathcal{G}, \mathcal{T}, \mathcal{S})
\end{equation}
where $\mathcal{G} = (V, E)$ represents the 3D molecular graph, $\mathcal{T} \in \mathbb{R}^{d_{tab}}$ is a vector of macroscopic physicochemical descriptors, and $\mathcal{S} \in \mathbb{R}^{d_{sem}}$ is the dense sequence embedding derived from a pretrained language model. 

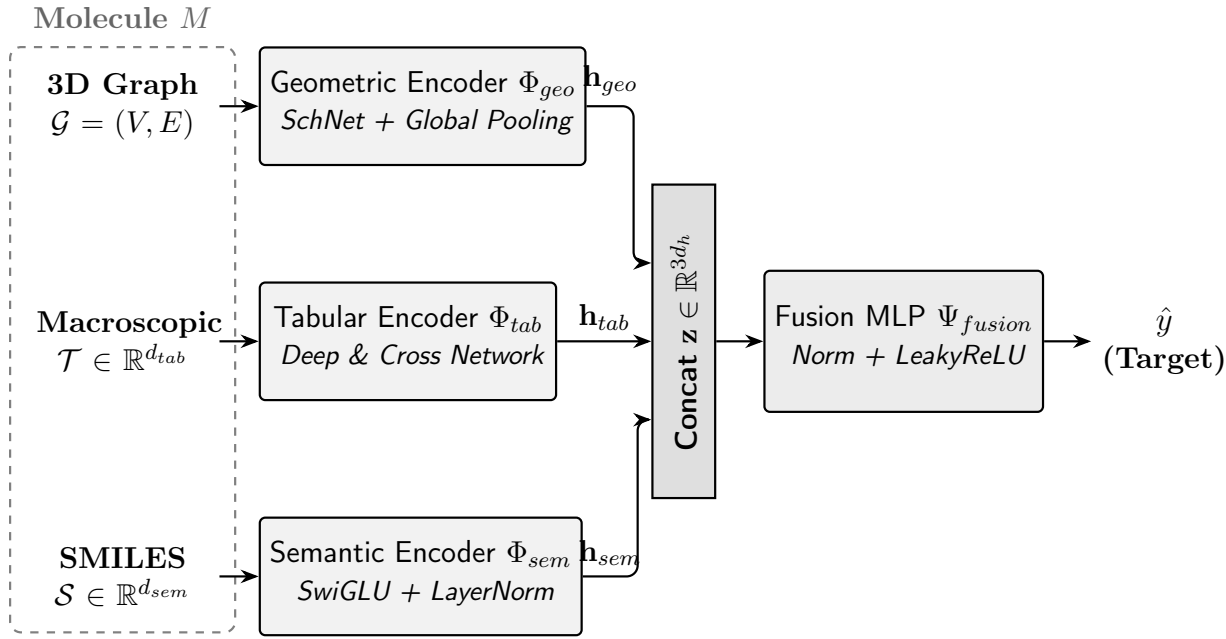
\begin{figure}[htbp]
	\centering
	\resizebox{\linewidth}{!}{
		\begin{tikzpicture}[
			font=\sffamily,
			>=Stealth,
			inputNode/.style={align=center, text width=2.2cm, font=\bfseries},
			encoderNode/.style={draw, rectangle, rounded corners=2pt, fill=gray!10, 
				minimum width=3.5cm, minimum height=1.5cm, align=center, thick},
			concatNode/.style={draw, rectangle, fill=gray!25, minimum width=0.8cm, 
				minimum height=4cm, align=center, thick},
			fusionNode/.style={draw, rectangle, rounded corners=2pt, fill=gray!15, 
				minimum width=2.8cm, minimum height=1.8cm, align=center, thick},
			outputNode/.style={align=center, font=\bfseries\large},
			arrow/.style={->, thick, rounded corners=4pt}
			]
			
			% --- 1. INPUTS ---
			\node[inputNode] (InGeo) at (0, 3) {3D Graph\\$\mathcal{G} = (V, E)$};
			\node[inputNode] (InTab) at (0, 0) {Macroscopic\\$\mathcal{T} \in \mathbb{R}^{d_{tab}}$};
			\node[inputNode] (InSem) at (0, -3) {SMILES\\$\mathcal{S} \in \mathbb{R}^{d_{sem}}$};
			
			% --- 2. ENCODERS ---
			\node[encoderNode, right=0.5cm of InGeo] (EncGeo) {Geometric Encoder $\Phi_{geo}$\\ \vspace{0.1cm} \small \textit{SchNet + Global Pooling}};
			\node[encoderNode, right=0.5cm of InTab] (EncTab) {Tabular Encoder $\Phi_{tab}$\\ \vspace{0.1cm} \small \textit{Deep \& Cross Network}};
			\node[encoderNode, right=0.5cm of InSem] (EncSem) {Semantic Encoder $\Phi_{sem}$\\ \vspace{0.1cm} \small \textit{SwiGLU + LayerNorm}};
			
			% --- 3. CONCATENATION ---
			\node[concatNode, right=1.2cm of EncTab] (Concat) {\rotatebox{90}{\textbf{Concat} $\mathbf{z} \in \mathbb{R}^{3d_h}$}};
			
			% --- 4. FUSION NETWORK ---
			\node[fusionNode, right=0.6cm of Concat] (Fusion) {Fusion MLP $\Psi_{fusion}$\\ \vspace{0.1cm} \small \textit{Norm + LeakyReLU}};
			
			% --- 5. OUTPUT ---
			\node[outputNode, right=0.6cm of Fusion] (Output) {$\hat{y}$\\ \normalsize (Target)};
			
			% --- 6. ROUTING / ARROWS ---
			\draw[arrow] (InGeo) -- (EncGeo);
			\draw[arrow] (InTab) -- (EncTab);
			\draw[arrow] (InSem) -- (EncSem);
			
			\draw[arrow] (EncGeo.east) -- node[above] {$\mathbf{h}_{geo}$} ++(0.6cm,0) |- ([yshift=1cm]Concat.west);
			\draw[arrow] (EncTab.east) -- node[above] {$\mathbf{h}_{tab}$} (Concat.west);
			\draw[arrow] (EncSem.east) -- node[above] {$\mathbf{h}_{sem}$} ++(0.7cm,0) |- ([yshift=-1cm]Concat.west);
			
			\draw[arrow] (Concat) -- (Fusion);
			\draw[arrow] (Fusion) -- (Output);
			
			% --- 7. MOLECULE GROUPING ---
			\begin{scope}[on background layer]
				\node[draw, dashed, thick, gray, rounded corners, fit=(InGeo) (InTab) (InSem), inner sep=0.2cm] (MoleculeBox) {};
				\node[above=0.1cm of MoleculeBox, font=\bfseries\color{gray!80!black}] {Molecule $M$};
			\end{scope}
			
		\end{tikzpicture}
	}
	\caption{\textbf{Schematic of the Tri-Branch Modular Fusion Network.} The framework generalizes molecular representation by integrating a continuous-filter GNN for 3D geometries, a DCN for explicit tabular physics, and a SwiGLU-gated encoder for SMILES sequences. The aligned latent representations ($\mathbf{h}_{geo}, \mathbf{h}_{tab}, \mathbf{h}_{sem}$) are concatenated and processed through a shared late-fusion network to predict the target extensive property ($\hat{y}$).}
	\label{fig:master_architecture}
\end{figure}

As illustrated in Figure \ref{fig:master_architecture}, these modalities are routed through three independent parameterized encoders—Geometric ($\Phi_{geo}$), Tabular ($\Phi_{tab}$), and Semantic ($\Phi_{sem}$)—to extract structurally aligned latent vectors of a uniform hidden dimension $d_h$. This parallel routing funnels the orthogonal representations into a single unified state before projection to the target property.

\subsection{Geometric Encoder ($\Phi_{geo}$)}
To capture the complex interatomic spatial dependencies required for quantum-level predictions, our geometric branch utilizes a continuous-filter convolutional architecture \cite{schutt2017schnet}. Each atom $i \in V$ is associated with an atomic number $Z_i$ and Cartesian coordinates $\mathbf{r}_i \in \mathbb{R}^3$. Interatomic distances $d_{ij} = \Vert \mathbf{r}_i - \mathbf{r}_j \Vert_2$ parameterize a learnable continuous filter network to update node representations through iterative message passing.

In this work, we select atomization energy at 0 K ($U_0^{\mathrm{atom}}$) as our primary predictive target to evaluate the framework. To structurally adapt the architecture to this specific target, the geometric readout mechanism must be explicitly configured to match its underlying physics. Because $U_0^{\mathrm{atom}}$ is an extensive thermodynamic property—meaning its total magnitude scales directly with the number of atoms in the molecular system—standard size-invariant readouts (such as mean or max pooling) would improperly collapse this critical size-dependent information. 

Therefore, after $L$ interaction layers, we bypass standard readouts. The raw spatial messages $\mathbf{h}_i^{(L)}$ are first refined via an atom-wise Multilayer Perceptron ($\mathrm{MLP}_{\mathrm{atom}}$) and then strictly aggregated using global additive pooling:
\begin{equation}
	\mathbf{h}_{geo} = \sum_{i=1}^{|V|} \mathrm{MLP}_{\mathrm{atom}} \big( \mathbf{h}_i^{(L)} \big)
\end{equation}
By employing this additive constraint, the magnitude of the geometric latent representation ($\mathbf{h}_{geo} \in \mathbb{R}^{d_h}$) scales proportionally with the atom count ($|V|$), ensuring the resulting embedding space remains mathematically and physically consistent with extensive macroscopic properties.

\subsection{Tabular Encoder ($\Phi_{tab}$)}
To process the global physicochemical descriptors ($\mathcal{T}$), we utilize a DCN \cite{wang2017deep}. Unlike traditional MLPs that learn feature interactions implicitly, the DCN explicitly computes bounded polynomial cross-feature dependencies, providing the network with exact macroscopic heuristics.

For the $l$-th cross layer, the interaction is formulated as:
\begin{equation}
	\mathbf{x}_{l+1} = \mathbf{x}_0 \odot (\mathbf{W}_l^{\top}\mathbf{x}_l) + \mathbf{b}_l + \mathbf{x}_l
\end{equation}
where $\mathbf{x}_0$ is the standardized input vector, and $\mathbf{W}_l, \mathbf{b}_l$ are learnable parameters. Concurrently, a parallel deep network processes the input to extract implicit abstract patterns ($\mathbf{x}_{\text{deep}} = \mathrm{MLP}_{\mathrm{tab}}(\mathbf{x}_0)$). The outputs are concatenated and linearly projected:
\begin{equation}
	\mathbf{h}_{tab} = \mathbf{W}_{\text{final}} [ \mathbf{x}_{L_c} \parallel \mathbf{x}_{\text{deep}} ] + \mathbf{b}_{\text{final}}
\end{equation}
where $L_c$ is the total number of cross layers, yielding the final tabular representation $\mathbf{h}_{tab} \in \mathbb{R}^{d_h}$.

\subsection{Semantic Encoder ($\Phi_{sem}$)}
SMILES strings offer a robust topological mapping of molecular structure. However, directly concatenating high-dimensional textual embeddings introduces feature redundancy and risks drowning out the lower-dimensional physical modalities. To efficiently distill this embedding into the target latent space, we apply a SwiGLU-gated projection mechanism \cite{shazeer2020glu}.

The semantic vector is processed through parallel linear projections and gated via a Sigmoid Linear Unit (SiLU) activation, followed by dropout and layer normalization:
\begin{equation}
	\mathbf{h}_{sem} = \mathrm{LayerNorm}\Big(\mathrm{Dropout}\big((\mathcal{S}\mathbf{W}_1) \odot \mathrm{SiLU}(\mathcal{S}\mathbf{W}_2)\big)\Big)
\end{equation}
where $\mathbf{W}_1, \mathbf{W}_2 \in \mathbb{R}^{d_{sem} \times d_e}$. The intermediate semantic bottleneck dimension ($d_e$) acts as an informational strangulation point, systematically filtering linguistic noise while preserving critical long-range topological dependencies before projecting to $d_h$.

\subsection{Multimodal Fusion and Optimization}
The latent representations from the three distinct branches are concatenated to form a unified multimodal state ($\mathbf{z} \in \mathbb{R}^{3d_h}$):
\begin{equation}
	\mathbf{z} = [ \mathbf{h}_{geo} \parallel \mathbf{h}_{tab} \parallel \mathbf{h}_{sem} ]
\end{equation}
This concatenated vector is processed through a deep fusion network ($\Psi_{fusion}$) utilizing Layer Normalization \cite{ba2016layer} and LeakyReLU activations to maintain gradient stability across the heterogeneous inputs, ultimately outputting the normalized prediction $\hat{y}_{\mathrm{norm}}$.

To ensure scale-invariant gradient descent, target atomization energy values ($U_0^{\mathrm{atom}}$) are strictly standardized using training set statistics ($\mu_y, \sigma_y$). The framework is optimized end-to-end via the L1 Loss function (Mean Absolute Error):
\begin{equation}
	\mathcal{L}_{\mathrm{L1}} = \frac{1}{N} \sum_{i=1}^{N} \Big| \Big(\frac{y_i - \mu_y}{\sigma_y}\Big) - \hat{y}_{\mathrm{norm}, i} \Big|
\end{equation}
During inference, predictions are unnormalized ($\hat{y} = \hat{y}_{\mathrm{norm}} \cdot \sigma_y + \mu_y$) to compute the absolute error in physical units (eV).

\section{Experimental Setup}

\subsection{Dataset and Feature Engineering}
To evaluate the predictive performance of our multimodal framework, experiments were conducted using the QM9 molecular benchmark. To ensure strict data alignment across the spatial, physicochemical, and semantic domains, molecules that failed to generate valid 3D conformations, produced errors during RDKit descriptor calculation, or could not be tokenized by the semantic pipeline were systematically excluded. This intersection protocol yielded a highly curated dataset of exactly 129,012 structurally aligned molecules, which was partitioned into training and validation subsets using a standardized 80\%:20\% split.

For the tabular encoder ($\Phi_{tab}$), a fixed vector of $d_{tab} = 18$ physicochemical properties is extracted for each molecule using the RDKit API. As detailed in Table~\ref{tab:descriptors}, this feature set spans elemental composition, bond topology, global size metrics, and macroscopic heuristics. All 18 descriptor features are standardized via $z$-score normalization calculated over the training partition to ensure stable gradient descent.

\begin{table}[htbp]
	\centering
	\caption{The 18 physicochemical and topological descriptors extracted via RDKit.}
	\label{tab:descriptors}
	\begin{tabular}{ll}
		\hline
		\textbf{Feature Category} & \textbf{Specific Descriptors ($d_{tab} = 18$)} \\ \hline
		Elemental Composition & Counts of C, N, O, F, H \\
		Bond Topology & Counts of single, double, triple, and aromatic bonds \\
		Global Size Metrics & Total atoms, heavy atoms, Molecular Weight (MolWt) \\
		Macroscopic Heuristics & LogP, TPSA, FractionCSP3 \\
		Reactivity \& Flexibility & H-Bond Donors, H-Bond Acceptors, Rotatable Bonds \\ \hline
	\end{tabular}
\end{table}

For the semantic encoder ($\Phi_{sem}$), tokenized SMILES strings are processed through \texttt{ChemBERTa-77M-MLM}, a pretrained RoBERTa-style foundation model. This architecture fixes the input semantic dimensionality to $d_{sem} = 384$, providing a contextualized sequence feature map for each molecule.

\vspace{0.5em}
\noindent \textbf{Prediction Target and Normalization:} 
In this work, the primary predictive target is the extensive atomization energy at 0 K ($U_0^{\mathrm{atom}}$). To ensure scale-invariant optimization, ground-truth targets are $z$-score normalized prior to training using global training set statistics ($\mu = -76.1823\text{ eV}, \sigma = 10.3195\text{ eV}$). All final Mean Absolute Error (MAE) evaluations are unnormalized and reported in physical units (eV).

\subsection{Implementation Details and Hyperparameters}
To structurally align the feature spaces prior to late fusion, the hidden capacity for the geometric, tabular, and semantic pathways is fixed to a uniform dimension of $d_h = 128$. This constraint prevents the high-dimensional semantic embeddings ($d_{sem} = 384$) from overpowering the multi-source gradient flow. Gradient stability across these heterogeneous modalities is further enforced via Layer Normalization (\texttt{LayerNorm}) at the bottlenecks of the tabular, semantic, and fusion networks.

The geometric branch utilizes $L=6$ interaction blocks with 50 Gaussian filters and a cutoff radius of 10.0 \AA. Crucially, to enable equitable late-stage multimodal fusion, the standard SchNet scalar readout mechanism is bypassed. Instead, the raw spatial messages are explicitly refined through a custom atom-wise MLP ($128 \rightarrow 64 \rightarrow 128$) with Softplus activations prior to global additive pooling. While truncating the native readout slightly restricts the absolute peak performance of the standalone geometric model compared to highly tuned, monolithic packages, it is a deliberate and necessary architectural control. It enforces strict latent parity ($d_h=128$) across all modalities, ensuring that the fusion head measures the true synergistic gains of multimodal integration rather than being artificially dominated by an over-parameterized spatial branch.

\begin{table}[htbp]
	\centering
	\caption{Detailed architectural hyperparameters and training configuration.}
	\label{tab:hyperparameters}
	\begin{tabular}{ll}
		\hline
		\textbf{Hyperparameter} & \textbf{Value/Configuration} \\ \hline
		\multicolumn{2}{l}{\textit{Optimization}} \\
		Optimizer & AdamW (Fused) \\
		Learning Rate & 3e-4 (\texttt{ReduceLROnPlateau}, Factor 0.5) \\
		Batch Size & 64 \\
		Loss Function & L1 Loss (Normalized Targets) \\
		Early Stopping & 100 Epochs Patience \\ \hline
		\multicolumn{2}{l}{\textit{Geometric Branch ($\Phi_{geo}$)}} \\
		Interaction Blocks & $L=6$ \\
		Gaussian Filters & 50 (Cutoff: 10.0 \AA) \\
		Atom-wise MLP & $128 \rightarrow 64 \rightarrow 128$ (\texttt{Softplus}) \\ \hline
		\multicolumn{2}{l}{\textit{Tabular Branch ($\Phi_{tab}$)}} \\
		Cross Layers & $L_c=3$ \\
		Deep MLP & $128 \rightarrow 128$ (\texttt{ReLU}, Dropout $0.2$) \\ \hline
		\multicolumn{2}{l}{\textit{Semantic Branch ($\Phi_{sem}$)}} \\
		Projection Dimension ($d_e$) & Swept $\{32, \mathbf{64}, 128, 256\}$ \\
		Activation \& Regularization & \texttt{SiLU}, Dropout $0.1$ \\ \hline
		\multicolumn{2}{l}{\textit{Late Fusion Network ($\Psi_{fusion}$)}} \\
		Hidden Architecture & $384 \rightarrow 256 \rightarrow 128 \rightarrow 1$ \\
		Activations & \texttt{LeakyReLU} \\ \hline
	\end{tabular}
\end{table}

\subsection{Evaluation Metrics and Ablation Design}
Generalization is evaluated on the validation partition using Mean Absolute Error (MAE) and the coefficient of determination ($R^2$). To guarantee a mathematically rigorous evaluation, all baseline and partial configurations are subjected to identical data splits, normalization parameters, and learning rate schedules as the proposed architecture. 

Rather than chasing absolute leaderboard dominance through parameter scaling, our primary objective is to investigate how explicitly distinct representations complement one another. To systematically quantify this synergy, we establish an exhaustive, three-phase ablation matrix:

\textbf{Phase 1: Combinatorial Synergy.} Using the standalone 3D graph neural network (\textit{SchNet Only}) as our controlled anchor baseline, we systematically activate, deactivate, and pair individual encoder pathways. For this phase, the semantic projection bottleneck is strictly fixed to $d_e = 128$ to ensure perfect latent parity prior to fusion. This isolates performance across seven configurations: all three unimodal baselines, all three dual-modal pairings, and the full tri-modal framework, directly measuring the orthogonal complementarity of the data sources.

\textbf{Phase 2: Semantic Bottleneck Sensitivity.} We hypothesize that directly projecting high-dimensional text embeddings into a dense latent space risks over-parameterizing the fusion head and introducing linguistic noise. Freezing the internal geometric and tabular dimensions, we sweep the target compressed semantic dimension across a discrete grid ($d_e \in \{32, 64, 128, 256\}$) to empirically identify the optimal information bottleneck required to distill global topology.

\textbf{Phase 3: Tabular Architectural Validation.} To verify that exact macroscopic heuristics require specialized mathematical routing, we isolate the tabular branch within the optimal tri-modal framework. We substitute the DCN cross-network with a standard Multi-Layer Perceptron (MLP) of equivalent parameter capacity. This isolates whether explicit polynomial feature crossing establishes a superior error floor compared to implicit dense mixing.

\section{Results and Discussion}

\subsection{Modality Ablation Analysis}
To systematically verify the core hypothesis of this work—that geometric structures, tabular chemical descriptors, and text-semantic sequences capture orthogonal, highly synergistic molecular features—we benchmark our multimodal configurations against the pure geometric baseline (\textit{SchNet Only}). For this initial combinatorial phase, the semantic bottleneck is fixed to $d_e = 128$ to maintain strict latent alignment with the spatial and tabular branches.

\begin{figure*}[htbp]
	\centering
	\includegraphics[width=\textwidth]{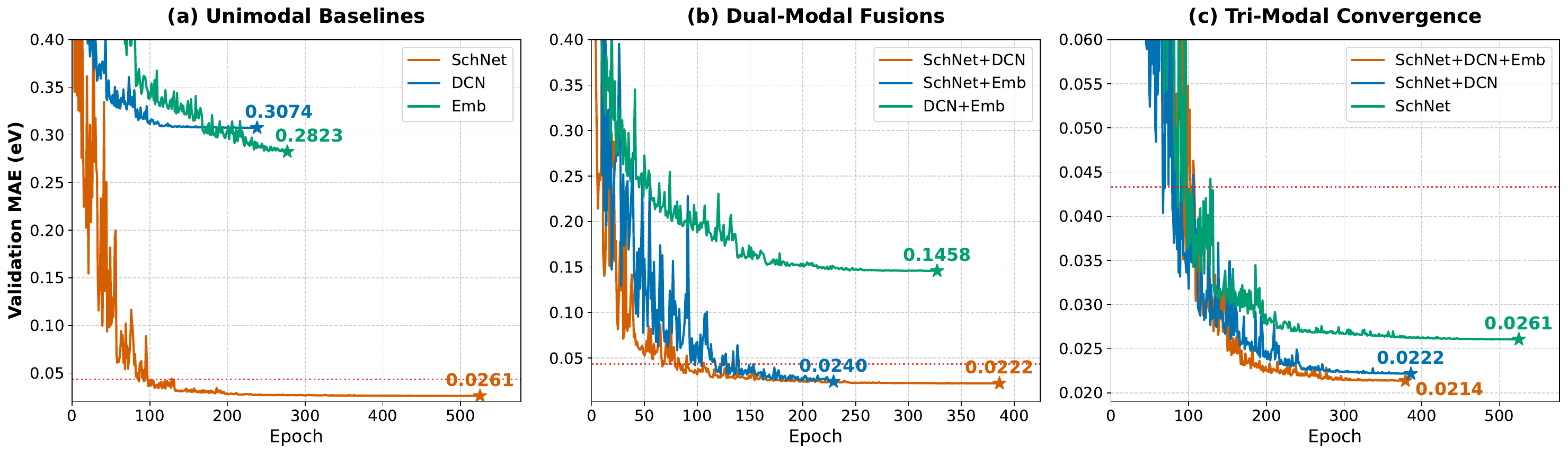}
	\caption{\textbf{Multimodal Synergy Ablation.} Validation MAE convergence profiles demonstrating systematic error reduction via modality fusion. Star markers ($\star$) indicate the optimally truncated checkpoint. The tri-modal convergence (Panel c) highlights both a significantly lower absolute error floor and accelerated, stabilized gradient dynamics compared to the isolated SchNet baseline.}
	\label{fig:synergy_1x3}
\end{figure*}

\begin{table}[htbp]
	\centering
	\caption{Combinatorial ablation performance evaluated at final convergence. For these baseline comparisons, the semantic projection dimension is structurally aligned to $d_e=128$.}
	\label{tab:master_ablation}
	\begin{tabular}{lccc}
		\hline
		\textbf{Model Configuration} & \textbf{Train Loss ($\downarrow$)} & \textbf{Val MAE (eV) $\downarrow$} & \textbf{Val $R^2$ ($\uparrow$)} \\ \hline
		\multicolumn{4}{l}{\textit{Unimodal Baselines}} \\
		DCN Only                     & 0.0263 & 0.3074 & 0.9973 \\
		Embedding Only               & 0.0224 & 0.2823 & 0.9977 \\
		\textbf{SchNet Only (Baseline)} & \textbf{0.0016} & \textbf{0.0261} & \textbf{0.9999} \\ \hline
		\multicolumn{4}{l}{\textit{Dual-Modal Synergies}} \\
		DCN + Embedding              & 0.0087 & 0.1458 & 0.9992 \\
		SchNet + Embedding           & 0.0014 & 0.0240 & 0.9999 \\
		SchNet + DCN                 & 0.0012 & 0.0222 & 0.9999 \\ \hline
		\multicolumn{4}{l}{\textit{Tri-Modal Framework}} \\
		\textbf{Full Framework ($d_e=128$)} & \textbf{0.0010} & \textbf{0.0214} & \textbf{0.9999} \\ \hline
	\end{tabular}
\end{table}

The experimental metrics, consolidated in Table~\ref{tab:master_ablation}, establish a rigorous performance hierarchy that exposes the underlying information dynamics between the data sources. 

\subsubsection{Unimodal Baselines and Contextual Constraints}
Among isolated modalities (Figure~\ref{fig:synergy_1x3}a), spatial coordination data predictably provides the strongest unimodal signal. The \textit{SchNet Only} baseline establishes a robust performance floor with a validation MAE of 0.0261 eV. 

It is important to acknowledge that this absolute baseline error trails the highly optimized, standalone SchNet implementations frequently reported in state-of-the-art literature. This discrepancy is a deliberate byproduct of our strict experimental controls designed to measure multimodal synergy. First, to enable equitable late-stage fusion, the native scalar readout mechanism was truncated to output a high-dimensional embedding ($d_h=128$), sacrificing the deep, specialized atom-wise energy networks typically employed in optimized packages (e.g., SchNetPack). Second, the training hyperparameters and learning rate schedules were globally standardized to ensure stable multi-source gradient flow across the tri-modal architecture, rather than being aggressively hyper-tuned for the spatial branch in isolation. Consequently, this baseline serves its intended purpose: acting as a strictly controlled anchor to isolate the \textit{relative} synergistic gains of multimodal integration, rather than competing with monolithic package-level optimizations.

When comparing the non-geometric baselines, the \textit{Embedding Only} model (0.2823 eV) outperforms the \textit{DCN Only} model (0.3074 eV). This gap is directly attributable to their pretraining paradigms: ChemBERTa captures structural syntax and functional group topology via masked language modeling, whereas the DCN relies on a highly compressed vector of predefined macroscopic scalars. Unsurprisingly, without 3D spatial coordinates, these isolated modalities cannot resolve complex quantum energy landscapes, further validating the necessity of the geometric core.

\subsubsection{Dual-Modal Synergies and Orthogonality}
The true impact of multimodal representation emerges when evaluating the dual-modal pairings. Pairing modalities induces a striking, non-linear error reduction driven by feature orthogonality (Figure~\ref{fig:synergy_1x3}b). 

The most prominent evidence of this synergy is observed in the \textit{DCN + Embedding} configuration. Despite completely lacking 3D atomic coordinates, fusing explicit physics (DCN) with global topology (SMILES) slashes the error nearly in half compared to their isolated states, achieving 0.1458 eV. This massive relative improvement strongly supports the hypothesis that macroscopic heuristics and topological sequences provide deeply complementary representations of the molecule.

Even more revealing is the synergistic reversal that occurs when secondary modalities are fused with the 3D geometric core. Although text outperformed tabular descriptors in isolation, \textit{SchNet + DCN} (0.0222 eV) yields a lower error than \textit{SchNet + Embedding} (0.0240 eV). Theoretically, exact macroscopic properties (e.g., atom counts, molecular weight) are implicitly encoded within the 3D graph. However, message-passing GNNs are notoriously inefficient at exact global counting tasks due to oversmoothing \cite{chen2020can, li2018deeper}. By injecting explicitly calculated descriptors directly into the fusion head, the tabular branch acts as a physical shortcut \cite{karniadakis2021physics}. This pairing covers the GNN's physical arithmetic blind spots, resulting in robust dual-modal synergy.

\subsubsection{The Tri-Modal Framework and Topological Scaling}
When all three data streams are synthesized, the model reaches its highest predictive performance (0.0214 eV for the 128-dimensional variant). As visualized in Figure~\ref{fig:synergy_1x3}c, the tri-modal architecture consistently outperforms all isolated and dual-modality baselines, establishing a significantly lower absolute error floor.

The mechanism driving this profound synergy lies in how the modalities structurally complement each other. An important architectural question naturally arises: if SchNet inherently maps the 3D graph, why does integrating a 1D SMILES sequence yield further performance gains rather than causing redundant feature collapse? 

The answer stems from the contrasting mathematical paradigms of their respective architectures. SchNet resolves local quantum micro-environments via message passing but struggles to efficiently exchange information between distant nodes. Conversely, the semantic transformer processes topology globally; via self-attention, the distance between any two tokens is effectively $\mathcal{O}(1)$ \cite{vaswani2017attention}. Ultimately, the geometric branch maps localized micro-physics, the tabular branch ensures exact macroscopic heuristics, and the semantic branch acts as a global topological grammar, seamlessly resolving the long-range dependencies that pure 3D convolutions often obscure.

\subsection{Semantic Bottleneck Evaluation}
While the tri-modal architecture yields superior predictive capabilities, projecting 384-dimensional text features directly into a 128-dimensional latent space risks sub-optimal feature mixing. To maximize fusion synergy, we evaluated the SwiGLU projection layer by freezing the geometric and tabular latent spaces at 128 and sweeping the semantic bottleneck dimension ($d_e$) across a discrete search grid. 

\begin{table}[htbp]
	\centering
	\caption{Impact of the semantic embedding bottleneck size ($d_e$) on final tri-modal optimization.}
	\label{tab:bottleneck_tuning}
	\begin{tabular}{cccc}
		\hline
		\textbf{Bottleneck Dimension ($d_e$)} & \textbf{Train Loss ($\downarrow$)} & \textbf{Val MAE (eV) $\downarrow$} & \textbf{Val $R^2$ ($\uparrow$)} \\ \hline
		32  & 0.0010 & 0.0216 & 0.9999 \\ 
		\textbf{64 (Optimal)}  & \textbf{0.0009} & \textbf{0.0207} & \textbf{0.9999} \\
		128 & 0.0010 & 0.0214 & 0.9999 \\ 
		256 & 0.0011 & 0.0220 & 0.9999 \\ \hline
	\end{tabular}
\end{table}

Configuring the semantic bottleneck to exactly $d_e = 64$ establishes the structural optimum, achieving the lowest validation MAE of 0.0207 eV (Table~\ref{tab:bottleneck_tuning}). This represents a \textbf{substantial 20.6\% error reduction} compared to the unimodal geometric baseline. The parabolic shape of the validation error confirms that synergistic fusion requires controlled feature compression. Scaling past the optimal threshold ($d_e=256$) over-parameterizes the projection, passing linguistic noise that drowns out the precise spatial coordinates. Restricting the space to exactly 64 dimensions forces the network to distill only the most generalizable topological semantics.

\begin{figure*}[htbp]
	\centering
	\includegraphics[width=\textwidth]{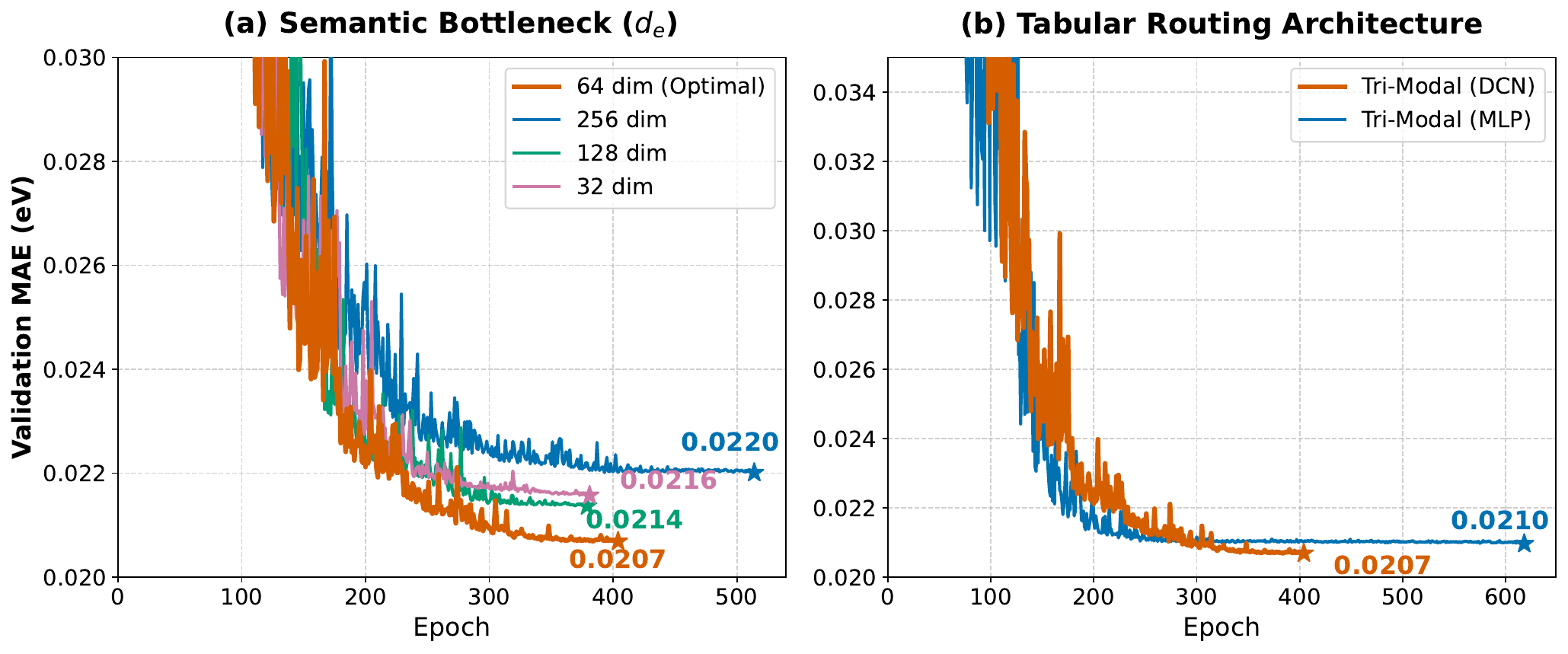}
	\caption{\textbf{Framework Optimization.} \textbf{(a)} Validation trajectories across varying semantic bottleneck dimensions ($d_e$). \textbf{(b)} Architectural ablation of the tabular branch. While explicit polynomial crossing (DCN) provides a marginal edge, the primary performance gain stems from the orthogonal macroscopic data itself.}
	\label{fig:optimizations}
\end{figure*}

\subsubsection{Architectural Validation of the Tabular Branch}
To rigorously evaluate whether the synergy is driven by the \textit{data} or the \textit{architecture}, we conducted a structural control experiment. Using the optimized tri-modal framework ($d_e = 64$), we substituted the explicit cross-network (DCN) with a standard Multi-Layer Perceptron (MLP) of equivalent parameter capacity. 

\begin{table}[htbp]
	\centering
	\caption{Comparison of tabular routing architectures within the optimal framework ($d_e=64$).}
	\label{tab:dcn_vs_mlp}
	\begin{tabular}{lccc}
		\hline
		\textbf{Tabular Architecture} & \textbf{Train Loss ($\downarrow$)} & \textbf{Val MAE (eV) $\downarrow$} & \textbf{Val $R^2$ ($\uparrow$)} \\ \hline
		Tri-Modal (MLP Branch) & 0.0013 & 0.0210 & 0.9999 \\ 
		\textbf{Tri-Modal (DCN Branch)} & \textbf{0.0009} & \textbf{0.0207} & \textbf{0.9999} \\ \hline
	\end{tabular}
\end{table}

The MLP control achieved a highly competitive validation MAE of 0.0210 eV, performing closely to the optimized DCN configuration (0.0207 eV). This marginal performance delta confirms our foundational theory. Standard MLPs are effective universal approximators for dense, low-dimensional continuous spaces \cite{gorishniy2021revisiting}. Because our tabular branch relies on a compact vector of only 18 continuous descriptors, an MLP can implicitly approximate the necessary interactions. 

Consequently, while the DCN maintains a mathematical edge by explicitly bounding polynomial interactions, the MLP control proves the overarching thesis of this framework: the profound performance gains over standard spatial networks are driven primarily by the synergistic inclusion of orthogonal data modalities, rather than relying exclusively on complex architectural depth.

\subsection{Limitations and Architectural Trade-offs}

While the proposed tri-modal architecture achieves a validation MAE of 0.0207 eV for atomization energy ($U_0^{\mathrm{atom}}$)—comfortably surpassing the sub-chemical accuracy threshold—several architectural constraints contextualize these findings and illuminate pathways for future refinement.

First, to conduct a strictly controlled multimodal ablation, we deliberately constrained the geometric baseline. The isolated \textit{SchNet Only} architecture was truncated via a custom atom-wise MLP to force a 128-dimensional latent bottleneck at the readout layer. Although this modification successfully enforced structural parity—preventing the spatial branch from artificially dominating the multi-source gradient flow—it restricted the standalone geometric model from reaching its theoretical peak performance. Substituting this backbone with a massively parameterized equivariant network (e.g., MACE or NequIP) would undoubtedly depress the absolute error floor further. However, such an upgrade would inherently sacrifice the sub-million parameter footprint and rapid inference capabilities that define the core utility of this framework.

Second, the semantic sequence branch utilizes a frozen, pretrained ChemBERTa language model. While freezing these 77 million parameters critically prevents the high-dimensional language model from overfitting and overpowering the lighter physical branches during early optimization, it relegates the transformer to a static feature extractor. Consequently, the semantic encoder cannot fine-tune its learned topological syntax specifically to the $U_0^{\mathrm{atom}}$ thermodynamic energy landscape. Future iterations could integrate Parameter-Efficient Fine-Tuning (PEFT) techniques, such as Low-Rank Adaptation (LoRA), to enable end-to-end semantic fine-tuning without violating the strict VRAM and parameter efficiency constraints of the architecture.

Third, the empirical optimization dynamics reported in this study—including the identified semantic bottleneck ($d_e = 64$) and the overall convergence trajectories—are inherently bounded by the defined hyperparameter search grid and available hardware compute. While our specific configuration establishes a mathematically robust proof-of-concept for multimodal synergy, a more exhaustive, compute-intensive Neural Architecture Search (NAS) across alternative fusion depths, learning rate schedules, and gating mechanisms could yield distinct optimization profiles. Scaling this hyperparameter search utilizing distributed compute clusters remains a necessary future step to identify the absolute global minimum of the tri-modal design space.

Finally, the empirical evaluation is bounded by the inherent scope of the QM9 benchmark, which is restricted to small, equilibrium-relaxed organic molecules (up to nine heavy atoms). As a result, the framework's capacity to maintain synergistic multimodal alignment when scaling to large macromolecules, highly flexible poly-pharmacological compounds, or non-equilibrium transitional states remains an open question. Future work will aim to extend this tri-modal synthesis to dynamic molecular dynamics (MD) trajectories and structurally diverse datasets (e.g., PCQM4Mv2) to rigorously evaluate its predictive stability across shifting conformational landscapes.

\section{Conclusion}

In this work, we introduced a parameter-efficient Tri-Branch Modular Fusion Neural Network that systematically addresses the representational limitations of isolated molecular modalities. By synthesizing continuous 3D spatial geometry (SchNet), discrete topological grammar (SMILES via ChemBERTa), and explicitly calculated macroscopic heuristics (Deep \& Cross Network), we established a highly orthogonal, multimodal latent space for robust molecular property prediction. 

Empirical evaluation on the QM9 dataset demonstrates the exceptional predictive fidelity of this multimodal approach. By optimizing the semantic projection bottleneck to $d_e=64$, the tri-modal framework achieved a validation MAE of $0.0207$~eV for atomization energy ($U_0^{\mathrm{atom}}$). Operating with fewer than one million parameters, this architecture not only decisively surpasses the standard sub-chemical accuracy threshold ($\approx 0.0433$~eV), but also achieves a substantial $20.6\%$ error reduction over a strictly controlled, pure geometric baseline. This extreme parameter efficiency enables rapid inference, positioning the framework as a highly competitive surrogate for large-scale, high-throughput virtual screening (HTVS) pipelines.

The success of this architecture is fundamentally grounded in information theory and physical complementarity. While modern message-passing graph neural networks excel at resolving continuous local micro-environments, they inherently struggle with exact global counting tasks and often suffer from oversmoothing across long-range dependencies. Our ablation studies demonstrate that injecting tabular heuristics acts as a vital physical shortcut, effectively bypassing the network's arithmetic bottlenecks. Furthermore, integrating transformer-based textual embeddings resolves macroscopic topology; by leveraging global self-attention, it anchors the localized 3D graph with a codified, $\mathcal{O}(1)$ semantic prior.

Looking forward, this multimodal alignment strategy provides a scalable blueprint for next-generation molecular representation. As the field gravitates toward massively over-parameterized monolithic geometric models, our findings offer an alternative, highly efficient paradigm: fusing spatial networks with orthogonal macroscopic and topological data streams can effectively break current performance asymptotes without brute-force parameter scaling. Future work will explore parameter-efficient fine-tuning (PEFT) of the semantic encoder and aim to apply this tri-modal framework to dynamic molecular trajectories, extending its utility from static equilibrium states to complex, non-equilibrium thermodynamic environments.

\section*{Acknowledgements}
The authors acknowledge the use of Google Gemini (version Gemini 1.5 Pro) during the preparation of this manuscript. Specifically, the AI was utilized as a writing assistant to polish the academic prose, ensure consistent mathematical formatting, and improve structural flow.

\end{document}